\documentclass[smallextended]{svjour3}       
\smartqed  
\usepackage{graphicx}
\usepackage{amssymb}
\usepackage{amsmath}
\usepackage{times}
\usepackage{epsfig}
\usepackage{float}
\usepackage{booktabs}
\usepackage{array}
\usepackage[ruled,vlined,titlenumbered]{algorithm2e}

\SetKw{nil}{nil}

\newcommand{\RR}{\ensuremath{\mathbb{R}}}

\newcommand{\normzero}[1]{{\|{#1}\|}_{0}}
\newcommand{\normone}[1]{{\|{#1}\|}_{1}}
\newcommand{\normtwo}[1]{{\|{#1}\|}_{2}}
\newcommand{\normP}[1]{{\|{#1}\|}_{p}}
\newcommand{\sign}[1]{\mbox{sign}(#1)}
\newcommand{\card}[1]{{\left|#1\right|}}
\newcommand{\bucy}{\tilde{y}}
\newcommand{\kmax}{\bar{k}}
\newcommand{\bucz}{\tilde{z}}

\newcommand{\calB}{\mathcal{B}}

\usepackage{mathtools}
\DeclarePairedDelimiter\ceil{\lceil}{\rceil}

\newcommand{\figsize}{5.9cm }
\newcommand{\buckP}{$bucket^F$ } 
\newcommand{\condP}{$pivot^F$ } 
\newcommand{\wbuckP}{$w$-$bucket^F$ } 
\newcommand{\wbuckPn}{$w$-$bucket$ } 
\newcommand{\wcondP}{$w$-$pivot^F$ }

\begin{document}


\title{Efficient Projection Algorithms onto the Weighted $\ell_1$ Ball}


\author{Guillaume Perez$^1$         \and
        Sebastian Ament$^2$         \and
        Carla Gomes$^2$             \and
        Michel Barlaud$^3$
}


\institute{
$^1$Huawei Technologies Ltd, French Research Center\\
$^2$Cornell University, Ithaca, New York, 14850, United States\\
$^3$Universit\'e C\^ote d'Azur, CNRS, Sophia Antipolis, 06900, 
\\ 
\email{ guillaume.perez06@gmail.com,\\
barlaud@i3s.unice.fr}
}

\date{Received: date / Accepted: date}

\maketitle

\begin{abstract}
Projected gradient descent has been proved efficient in many optimization and machine learning problems. 
The weighted $\ell_1$ ball has been shown effective in sparse system identification and features selection.
In this paper we propose three new efficient algorithms for projecting any vector of finite length onto the weighted $\ell_1$ ball.
The first two algorithms have a linear worst case complexity.  The third one has a  highly competitive performances in practice but the worst case has a quadratic complexity.
These new algorithms are efficient tools for machine learning methods based on projected gradient descent such as compress sensing, feature selection. 
We illustrate this effectiveness by adapting an efficient compress sensing algorithm to weighted projections.
We demonstrate the efficiency of our new algorithms on benchmarks using very large vectors. For instance, it requires only 8 ms, on an Intel I7 3rd generation, for projecting vectors of size $10^7$.
\end{abstract}

\section{Introduction}
Looking for sparsity appears in many machine learning applications, such as biomarker identification in biology \cite{abeel2009robust,he2010stable}, or the recovery of sparse signals in compressed sensing \cite{donoho2006compressed}.
For example consider the problem of minimizing a common $\ell_2$ reconstruction loss function. 
In addition consider constraining the number of non-zero components ($\ell_0$ norm) of the learned vector to be lower than a given sparsity value:
\begin{equation*}
\underset{x \in \RR^d}{\text{minimize}} \normtwo{Ax-b} \quad \text{ subject to }  \quad \normzero{x} < \epsilon
\end{equation*}
A solution of this generic problem will use only a subset of size lower than $\epsilon$ of the components of $x$ leading to the best reconstruction error.
This implies that solving this problem and varying the $\epsilon$ value allows us to manage sparsity with a fine grain.
Unfortunately, this problem is generally strictly nonconvex and very difficult to solve \cite{natarajan1995sparse}. 
Hence a common solution is to constrain the $\ell_1$ norm of the vector instead \cite{tibshirani96,donoho03}, or one of its modified versions \cite{slavakis10,kopsinis2010online,bogdan2015slope}.
More specifically, the weighted $\ell_1$ ball has been proven to have strong properties for finding or recovering sparse vectors \cite{candes2008enhancing,slavakis10}.
To this purpose, it is crucially important to have simple algorithms to work with the weighted $\ell_1$ ball during the optimization process. 
Several methods in machine learning are based on projected gradient descent for their optimization part, for example by iterating between gradient updates and $\ell_1$ ball projections \cite{duchi}. 
The goal of this paper is to help the application of projected gradient descent methods for solving sparse machine learning problems by working with the weighted $\ell_1$ ball. 
Unfortunately, efficiently applying projected gradient descent requires an efficient projection algorithm.
For the basic $\ell_1$ ball, many projection algorithms have been proposed \cite{michelot86,van08,kiwiel08,condat,perez19}, but in the context of the weighted $\ell_1$ ball, only few works have been done \cite{slavakis10}.

In this paper, we propose three efficient projection algorithms by generalizing works made for the basic (i.e. non weighted) $\ell_1$ ball to its generalization, the weighted $\ell_1$ ball.
We start by giving a proof of existence of the threshold value ($\lambda*$) allowing efficient projection algorithms as for the basic $\ell_1$ ball \cite{held74}. 
Using this threshold, we propose three efficient algorithms. 
First, \wcondP iteratively approximates $\lambda*$ using a pivot-based algorithm. 
It Splits the vector in two sub-vectors at each iteration and has quadratic worst case complexity, but is near linear in practice. 
Second, \wbuckPn and \wbuckP, two algorithms based on a bucket decomposition of the vector that efficiently detects components that will be zero in the projection. These two ones have linear worst case complexity.
We propose an experimental protocol using randomly generated high dimensional ($> 10^5$) vectors to show the performances of our algorithm. 
We then adapt the sparse vector recovery framework from \cite{trzasko08} to the proposed projection algorithms and show the efficiency and simplicity of the model.
%
The paper starts with the definitions of the projection onto the $\ell_1$ and weighted $\ell_1$ balls, and highlights the need of finding the $\lambda*$ value. 
In the projection section, the algorithm \wcondP is first defined and the algorithms \wbuckPn and \wbuckP are later defined. 
In the experiments section, an evaluation of the time performances shows that the proposed algorithms are order of magnitude faster than current projection methods, and then that they can be used to adapt existing machine-learning frameworks.
Finally, the proofs required for the proposed algorithms are given and followed by the conclusion.

\section{Definitions of the Projections}
\paragraph{$\ell_1$ ball} 
Given a vector $y=(y_1,y_2,\ldots,y_d)\in \RR^d$ and a real $a>0$, we aim at computing its projection $P_{\calB_a}(y)$ onto the $\ell_1$ ball $\calB_a$ of radius $a$:
\begin{equation}
\calB_a=\left\{x\in\RR^d| \;\;\normone{x}\leq a\right\},
\end{equation}
where $\normone{x}=\sum_{i=1}^{d}|x_i|$.
The projection $P_{\calB_a}(y)$ is defined by
\begin{equation}
P_{\calB_a}(y)=\arg \min\limits_{x \in \calB_a} \normtwo{x-y}
\end{equation}
where $\normtwo{x}$ is the Euclidean norm. 
As shown in \cite{duchi} and revisited in \cite{condat}, the projection onto the $\ell_1$ ball can be derived from the projection onto the simplex $\Delta_a$:
\begin{equation}
\Delta_a{=}\left\{x\in\RR^d\, | \sum_{i=1}^{d}x_i= a\;\mbox{and}\;x_i\geq 0,\forall i=1,\ldots,d\right\}\!.
\end{equation}
Let the sign function $\sign{v}$ defined as $\sign{v} = 1$ if $v > 0$, $\sign{v} = -1$ if $v<0$ and $\sign{v} = 0$ otherwise, for any real value $v\in\RR$. The projection of $y$ onto the $\ell_1$ ball is given by the following formula:
\begin{equation}
P_{\calB_a}(y) = \left\{
	\begin{array}{lll}
	y&\mbox{if}\;\; y\in\calB_a,& \\
	(\sign{y_1}x_1,\ldots,\sign{y_d}x_d)&\mbox{otherwise},&
	\end{array}\right.
\end{equation}
where $x=P_{\Delta_a}(|y|)$ with $|y| = (|y_1|,|y_2|...,|y_d|)$ is the projection of $|y|$ onto $\Delta_a$. 
An important property has been established to compute this projection. 
It was shown \cite{held74} that there exists a unique $\tau=\tau_y\in\RR$ such that
\begin{equation}\label{eq:tau}
x_i=\max\{y_i-\tau,0\}, \forall i=1,\ldots,d.
\end{equation}
The projection is almost equivalent to a thresholding operation. 
The main difficulty is to compute quickly the threshold $\tau_y$ for any vector $y$. Let $y_{(i)}$ be the $i$th largest value of $y$ such that $y_{(d)}\leq y_{(d-1)} \leq \ldots \leq y_{(1)}$. It is interesting to note that (\ref{eq:tau}) involves that $\sum_{i=i}^{d}\max\{y_i-\tau,0\}=a$. Let $S^*$ be the support of $x$, i.e., $S^*=\{i|x_i>0\}$. Then, 
$$
a=\sum_{i=1}^{d}x_i=\sum_{i\in S^*}x_i=\sum_{i\in S^*}(y_i-\tau).
$$
It follows that $\tau_y=(\sum_{i\in S^*}y_i-a)/\card{S^*}$ where $\card{S^*}$ is the number of elements of $S^*$. The following property allows us to compute the threshold $\tau_y$. Let
\begin{equation}\label{eq:varrhok}
\varrho_j(y)=\left(\sum_{i=1}^{j}y_{(i)}-a\right)/j
\end{equation}
for any $j=1,\ldots,d$. Then, it was shown that $\tau_y=\varrho_{K_y}(y)$ where 
\begin{equation}\label{eq:K}
K_y=\max\{k\in\{1,\ldots,d\}\;|\;\varrho_k(y)<y_{(k)}\}.
\end{equation}
Looking for $K_y$, or equivalently $y_{(K_y)}$,  allows us to find immediately the threshold $\tau_y$.
The most famous algorithm to compute the projection, which has been presented in \cite{held74}, is based on (\ref{eq:K}). It consists in sorting the values and then finding the first value satisfying (\ref{eq:K}). A possible implementation is given in Algorithm \ref{a:sort}. The worst case complexity of this algorithm is $O(d\log d)$. Several other methods have been proposed \cite{michelot86,van08,kiwiel08,condat,perez19}, outperforming this simple approach.

\begin{algorithm}[t]
\KwData{$y, a$}
$u \leftarrow sort(y)$\\
$K \leftarrow \max_{1\leq k \leq d} \{ k | (\sum_{r=1}^{k}{u_r}-a)/k < u_k\}$\\
$\tau \leftarrow (\sum_{r=1}^{K}{u_r}-a)/K$\\
\For{$i \in 1..d$}{
	$x_i \leftarrow \max(y_i - \tau, 0)$
}
\caption{Sort based algorithm \cite{held74}\label{a:sort}}
\end{algorithm}

\paragraph{Weighted $\ell_1$ ball}
Given a vector $y=(y_1,y_2,\ldots,y_d)\in \RR^d$, a vector $w=(w_1,w_2,\ldots,w_d)\in \RR^d$, $w_i > 0$
\footnote{We can consider $w_i>0$ instead of $w_i \geq 0$ without loss of generality since the associated entries of y will be present in the projection, and do not influence the processing of the rest of the projection.}
for all $i$, and a real $a>0$, we aim at computing its projection $P_{\calB_{w,a}}(y)$ onto the $w$ weighted $\ell_1$ ball $\calB_{w,a}$ of radius $a$:
\begin{equation}
\calB_{w,a}=\left\{x\in\RR^d| \;\;\sum_{i=1}^{d} w_i|x_i| \leq a\right\},
\end{equation}

The projection $P_{\calB_{w,a}}(y)$ is defined by
\begin{equation}
P_{\calB_{w,a}}(y)=\arg \min\limits_{x \in \calB_{w,a}} \normtwo{x-y}
\end{equation}
where $\normtwo{x}$ is the Euclidean norm.

As for the classical $\ell_1$ ball, the weighted projection operator can be derived from the weighted simplex $\Delta_{w,a}$ projection \cite{slavakis10}:
\begin{equation}
\Delta_{w,a}{=}\left\{x\in\RR^d\, | \sum_{i=1}^{d}w_i x_i= a\;\mbox{and}\;x_i\geq 0,\forall i=1,\ldots,d\right\}\!.
\end{equation}

The projection of $y$ onto the weighted $\ell_1$ ball is given by the following formula:
\begin{equation}
P_{\calB_{w,a}}(y) = \left\{
	\begin{array}{lll}
	y&\mbox{if}\;\; y\in\calB_{w,a},& \\
	(\sign{y_1}x_1,\ldots,\sign{y_d}x_d)&\mbox{otherwise},&
	\end{array}\right.
\end{equation}
where $x=P_{\Delta_{w,a}}(|y|)$ with $|y| = (|y_1|,|y_2|...,|y_d|)$ is the projection of $|y|$ onto $\Delta_{w,a}$. 
Once again, the fast computation of the projection $x=P_{\Delta_{w,a}}(y)$ for any vector $y$ is of utmost importance. 

Let the vector $y=(y_1,y_2,\ldots,y_d)\in \RR^d$, the vector $w=(w_1,w_2,\ldots,w_d)\in \RR^d$, $w_i \geq 0$ for all $i$, and a real $a>0$, if $y \not\in \Delta_{w,a}$, then there exists a unique $\lambda=\lambda_y\in\RR$ such that
\begin{equation}\label{eq:lambda}
    x_i = \max\{y_i-w_i \lambda,0\}, \forall i=1,\ldots,d.
\end{equation}

%
The proof is given in the proof section of this paper, and is used to derive three projection algorithms. 
These algorithms are generalizations of the current state-of-the-art algorithms for $\ell_1$ ball projection \cite{condat,perez19}.
The main difficulty is to compute quickly the threshold $\lambda_y$ for any vector $y$. 
Let $z \in \RR^d$ be the vector such that 
\begin{equation}\label{eq:zDef}
    z_i = \frac{y_i}{w_i}, \forall i=1,\ldots,d.
\end{equation}
Let $z_{(i)}$ be the $i$th largest value of $z$ such that $z_{(d)}\leq z_{(d-1)} \leq \ldots \leq z_{(1)}$. 
Let $y_{(i)}$ (resp. $w_{(i)}$) be the $j$th entry of $y$ such that $z_{(i)} = y_j/w_j$.
$y_{()}$ is a permutation of $y$ with respect to the order of $z_{()}$.
It is interesting to note that equation (\ref{eq:lambda}) implies that 
$\sum_{i=i}^{d}\max\{y_i-w_i \lambda,0\}=a$.
Let $S^*$ be the support of $x$, i.e., $S^*=\{i|x_i>0\}$. Then, 
$$
a=\sum_{i=1}^{d} w_ix_i = \sum_{i\in S^*}w_ix_i = \sum_{i\in S^*}w_i(y_i-w_i\lambda).
$$
It follows that 
\begin{equation}\label{eq:lambda_def}
\lambda=\frac{\sum_{i\in S^*}w_iy_i-a}{\sum_{i\in S^*}w_i^2}
\end{equation}
The following property compute the threshold $\lambda_j$. 
\begin{equation}\label{eq:varrhok-w}
\varrho_j(w,y)=\frac{\sum_{i=1}^j w_{(i)}y_{(i)}-a}{\sum_{i=1}^jw_{(i)}^2}
\end{equation}
for any $j=1,\ldots,d$.
Then, we have shown in the proof section that $\lambda_y=\varrho_{K_y}(w,y)$ where 
\begin{equation}\label{eq:K-w}
K_y=\max\{k\in\{1,\ldots,d\}\;|\;\varrho_k(w,y)<z_{(k)}\}.
\end{equation}
Looking for $K_y$, or equivalently $y_{(K_y)}$,  gives us immediately the threshold $\lambda$.

A direct algorithm to compute this projection, which is a generalization of \cite{held74}, is based on (\ref{eq:K-w}), and is given in Algorithm \ref{a:weighted}. 
This algorithm starts by sorting the values according to $z$, and then searches for the $K_y$ index. 
Note that once $z$ sorted, finding $K_y$ is easily done by starting from the largest value.
That is why Algorithm \ref{a:weighted} does not need more steps as in \cite{slavakis10,kopsinis2010online}.

\begin{algorithm}[t]
\KwData{$y, w, a$}
\textbf{Output}: $x = P_{\calB_{w,a}}(y)$\\
$z^u \leftarrow \{\frac{y_i}{w_i}| \forall i \in [1,d]\}$\\
$_{()}\leftarrow$ Permutation$\uparrow(z^u)$\\
$z \leftarrow \{z^u_{(i)} |\forall i \in [1,d] \}$\\
$J \leftarrow \max_{1\leq J \leq d} \{ \arg\max j : \frac{-a+\sum_{i=j+1}^d w_{(i)} y_{(i)}}{\sum_{i=j+1}^d w_{(i)}^2} > z_j \}$\\
$\lambda^* \leftarrow \frac{-a + \sum_{j=J+1}^d w_{(j)} y_{(j)}}{\sum_{j=J+1}^d w_{(j)}^2}$\\
\For{$i \in 1..d$}{
	$x_i \leftarrow $\ sign$(y_i)\max(y_i - w_i\lambda^*, 0)$
}
\caption{Weighted generalization of the sort based algorithm\label{a:weighted}.}
\end{algorithm}

\section{Efficient Projection}
Many works have focused on designing efficient algorithms for finding $P_{\calB_a}(y)$ given $y\in\RR^d$ and $a\in\RR$ \cite{michelot86,kiwiel08,van08,condat,perez19}. 
In this section, given $y$, $w$, and $a$, we propose to generalize some of these efficient algorithms to find the projection onto the weighted $\ell_1$ ball $\calB_{a,w}$.
Specifically, we generalize the methods from \cite{condat,perez19} that we respectively name \condP and \buckP.
Both of these methods are looking for the $\tau$ value such that the projection $P_{\calB_{a}}(y)=x$ is defined by equation (\ref{eq:tau}). 

For the weighted ball, we are looking for the $\lambda$ value such that the projection $P_{\calB_{a,w}}(y)=x$ is given by equation (\ref{eq:lambda}).
As for the basic $\ell_1$ ball, the weighted $\ell_1$ ball algorithms are looking for $K$ and $y_{(K)}$ from equation (\ref{eq:K-w}).
Let $z=\frac{y}{w}$. If $z$ is sorted in increasing order, then finding $\lambda$ can be easily done by iteratively processing $\varrho_i(w,y)$, with $i = d,...,1$ until $\varrho_i(w,y) \geq (z_i)$, as shown in Algorithm \ref{a:weighted}. 
A projection algorithm using as a first iteration a sort has already been proposed \cite{slavakis10}.
But most of the time $z$ is not sorted, and sorting $z$ is the exact operation that we want to avoid because of its time complexity, and that often, we are looking for sparse solutions, thus only a subset of the values of $y$ will remain relevant. 
This section is split into two parts. 
The first one is a generalization of the algorithm \condP \cite{condat}.
The second one is a generalization of the algorithm \buckP \cite{perez19}.

\subsection{\wcondP Algorithm}
In this section we propose a generalization of the \condP algorithm. The proposed algorithm is composed of three points that we name pivot, lower bound extraction and online filtering, and are described in the next paragraphs. 
The idea of the algorithm is the following, at each iteration, the vector is split into two sub-vectors, using a pivot value, and determine which sub-vector contains $y_{(K)}$ (pivot part).
%
In the mean time, a fine grain lower bound is defined as a pivot for an efficient splitting (lower bound part). 
Finally, the algorithm discards \textit{on the fly} values that are provably not part of $S^*$ (online filtering part).

%
\textbf{Pivot} It is often considered that for regular amount of data, the \textit{quicksort} algorithm is the fastest sort \cite{thomas2001}. 
The \textit{quicksort} algorithm splits the data into two partitions by using a pivot value. 
Values smaller than the pivot go into the first partition, others in the second.
The process is then applied recursively to both partitions.
In the context of the basic simplex projection, using a \textit{pivot like} algorithm led to some of the most efficient algorithms \cite{kiwiel08,duchi,michelot86,condat}. 
%
From equation (\ref{eq:lambda_def}), we can notice that the $\lambda$ value only requires the knowledge about elements of the set $S*$, but not that this set is ordered. 
Such a remark gives a hint in why partitioning instead of sorting could be beneficial.
Consider $p\in [z_{(1)},z_{(n)}]$, note that to get $z_{(1)}$ and $z_{(n)}$ only one pass over the vector $z$ is required.
One can partition $z$ into $z_{low}$ and $z_{high}$ by putting the elements smaller (resp larger) than $p$ from $z$. 
Consider that the size of $z_{high}$ is $j$ (i.e. it contains $j$ entries), then we can easily compute $\varrho_j(w,y)$. 
If $\varrho_j(w,y) \geq p$, then $\lambda \geq p$. 
If this condition is true, this implies that we can stop the processing of $z_{low}$ because $z_{(K_y)} \in z_{high}$. 
If the condition is false, if $\varrho_j(w,y) \leq p$, then $\lambda \leq p$, then we know that $\{i | z_{i} \in z_{high}\} \subseteq S* $. 
Using this knowledge, we can continue the processing of $z_{low}$.
%
%
%

%
\textbf{Lower bound as pivot} The choice for the pivot is of utmost importance for the global running time, we can easily show that the worst case complexity is $O(d^2)$. 
As for the basic simplex, one could seek for the median pivot \cite{kiwiel08}. 
From a complexity point of view, this could be an improvement, but our goal is not to partition equitably the data, but to efficiently find $\lambda$.
Instead, we will seek for a pivot which is a lower bound of $\lambda$. 

Let $V$ be any sub-sequence of $y$. Then the value 
\begin{equation}
    p_V = \frac{-a + \sum_{i\in V} w_{(i)} y_{(i)}}{\sum_{i\in V} w_{(i)}^2} \label{eq:lowerBA}
\end{equation}
is a lower bound of $\lambda$.
The proof can be found at the end of this paper.

Using $p_V$ as a pivot implies that elements of $z_{low}$ can always be discarded. 

\textbf{Proof}. If we use $p_V$ as a threshold value, and by definition of a lower bound, we have:
\begin{equation}
    a \leq  \sum_{i\in 1..d} w_{(i)} \max(y_{(i)} - w_{(i)} p_V,0) 
\end{equation}
If $z_i \in z_{low}$, it implies that $\max(y_{i} - w_{i} p_V,0)= 0$, which also implies that $\max(y_{i} - w_{i} \lambda,0)= 0$. 
Consider the algorithm iterating between the following step: 
Step 1) Set $p=p_{z_{high}}$. 
Step 2) remove elements of $z_{high}$ which are smaller than $p$. 
This algorithm will converge to a state where no element can be removed anymore.
This state implies that the resulting vector $z_{high}$ is $S*$.
%

%
\textbf{Online Filtering} Another optimization of this algorithm is the following, the pivot value does not need to wait until the end of step 2) before being updated, but can be updated after every element of $z_{high}$ is read.
Such an interactive update requires us to divide the algorithm in two parts.

The first part is the first pass over $y$, where we do not have any knowledge about $y$.
Let initialize $V=\{1\}$ and the pivot by $p_V=\frac{w_1 y_1-a}{w_1^2}$. 
Consider that we are processing the $j$th element of $y$, which implies that we have already processed all the elements in [$1,j-1$].
From these elements, we have built a sub-sequence $V$, and an associated pivot $p_V$.
If we have $z_i \leq p_V$, then, as before, we can discard this element.
Otherwise, if $z_i > p_V$, then we can add $i$ to $V$ because $z_i$ is potentially larger than $\lambda$. 
Once $V$ is updated, we can update $p_V$ incrementally without waiting for the pass to be over, and then process the $(j+1)$th element of $y$.

For the second part, we have hopefully already discarded several elements of $y$, and more importantly, we have a set $V$ containing elements that are potentially larger than $\lambda$. In addition, we have the associated pivot $p_V$, which is processed with respect to all the elements in $V$.
Consider that we pass over the elements of $z_i \in V$, if $z_i \leq p_V$, then we can remove $z_i$ from $V$, and update $p_V$ accordingly.
Note that just like for the first part, this can be done incrementally. 
When no more element can be removed from $V$, the algorithm finished and $V = S*$. A possible implementation is given in Algorithm \ref{a:wcond}.

In the same fashion as \cite{condat}, we have incorporated a refinement. 
During the first iteration, while processing the $i$th element, when the current pivot value is smaller than the one defined by the current value $p'=\frac{w_i y_i-a}{w_i^2}$, then $p'$ will become the new basis.
To do that, an additional set $v'$ is required and a cleanup step which ensures that $V$ contains all the elements $z_i$ larger than $p_V$ before the second part of the algorithm.
Note that the proposed algorithm works on the permuted space of $z$ without using $z$ for the calculation of $J$ and $\lambda$, which is a major difference with the non-weighted version.


\begin{algorithm}[t]
\normalsize
\KwData{$y, w, a$}
$v \leftarrow \{y_1\}$ \\
$\tilde{v} \leftarrow \emptyset$ \\
$\lambda' \leftarrow \frac{w_1 y_1-a}{w_1^2}$ \\
\For{$n \in 2..d$}{
	\If{$\frac{y_n}{w_n} > \lambda'$}{
	    $\lambda' \leftarrow \frac{w_n y_n - a + \sum_{i\in v} w_{(i)} y_{(i)}}{w_n^2 + \sum_{i\in v} w_{(i)}^2}$ \\
	    \uIf{$ \frac{w_n y_n-a}{w_1^2} < \lambda'$}{
	        Add $n$ to $v$ 
	    }\Else{
	        Add $v$ to $\tilde{v}$ \\
	        $v \leftarrow \{y_n\}$ \\
            $\lambda' \leftarrow \frac{w_n y_n-a}{w_n^2}$
	    }
	}
}
\If{$\tilde{v} \neq \emptyset$}{
    \For{$n \in \tilde{v} $}{
        \If{$\frac{y_n}{w_n} > \lambda'$}{
            Add $n$ to $v$ \\
            $\lambda' \leftarrow \frac{-a + \sum_{i\in v} w_{(i)} y_{(i)}}{\sum_{i\in v} w_{(i)}^2}$
        }
    }
}
\While{$|v|$ changes}{
    \For{$n \in v $}{
        \If{$\frac{y_n}{w_n} < \lambda'$}{
            Remove $n$ to $v$ \\
            $\lambda' \leftarrow \frac{-a + \sum_{i\in v} w_{(i)} y_{(i)}}{\sum_{i\in v} w_{(i)}^2}$
        }
    }
}

$\lambda* \leftarrow \lambda'$\\
\For{$i \in 1..|y|$}{
	$x_i \leftarrow \max(y_i - w_i\lambda*, 0)$
}
\caption{\wcondP\label{a:wcond}}
\end{algorithm}

\subsection{\wbuckP Algorithm}
In this section, we present the \wbuckP algorithm, which is a generalization of the linear time simplex projection \buckP \cite{perez19}. 
\wbuckP fundamental idea is to recursively split vector $z$ into $B \geq 2$ ordered sub-vectors (say buckets) $\bucz_b^k$ with $b=1,\ldots,B$ and $k=1,\ldots,\kmax$, while looking for $z_{(K)}$. 
We say that the sub-vectors are ordered in the sense that all elements of $\bucz_b^k$ are smaller than the ones of $\bucz_{b+1}^k$ for all $b=1,\ldots,B-1$.
The depth, or number of recursive splitting is $\kmax$.
In the description of \wcondP, from the two sub-vectors $z_{low}$ and $z_{high}$, only one of them was re-used at the next iteration. In the \wbuckP algorithm, only one of the $B$ buckets will be re-used at the next iteration. 
Such a fine grain gives us three possible states for the buckets, $<z_{(K)}$, $>z_{(K)}$, $?z_{(K)}$. Only one bucket will be in the uncertainty state ($?z_{(K)}$), this is the bucket that will be re-used. 

We define here the different components of the \wbuckP algorithm.
For any level $k+1 \geq 1$, consider the interval $I^{k+1}$ defined by
\begin{equation} \label{eq:Ibound}
I^{k+1}=[\min \bucz_{b_k}^k,\max \bucz_{b_k}^k]
\end{equation}
with $\min \bucz_b^k$ (resp. $\max \bucz_b^k$) the minimum (resp. maximum) element of sub-vector $\bucz_b^k$.

Consider a partition of $I^{k+1}$ into $B$ ordered sub-intervals $I_1^{k+1}$,\ldots, $I_B^{k+1}$. 
Let $h^{k+1}:I^{k+1}\mapsto \{1,\ldots,B\}$ be the bucketing function such that $h^{k+1}(v)=b$ when the real value $v$ belongs to $I_b^{k+1}$.
The bucket $\bucz_{b_k}^{k}$ is split into $B$ ordered sub-vectors $\bucz_{b}^{k+1}$ such that
\begin{enumerate}
    \item $S_b^{k+1}=\{i\in S_{b_k}^{k}\;:\;h^{k+1}(z_i)=b\}$,
    \item $\bucz_b^{k+1}=(z_i)_{i\in S_b^{k+1}}$,
   
    \item $\max \bucz_b^{k+1}< \min \bucz_{b+1}^{k+1}$ for all $b=1,\ldots,B-1$,
\end{enumerate}
with the convention $S_{b_0}^{0}=\{1,\ldots,d\}$. We get $\card{S_{B}^{k}}\geq 1$ at any level $k\geq 1$ because of the definition of $I^{k+1}$. The fact that $\max \bucz_b^{k+1}< \min \bucz_{b+1}^{k+1}$ follows from the fact that equal values of $z$ necessarily belongs to the same bucket. 

For any $k>0$, 
\begin{equation}\label{eq:cumsum}
C_{b}^{k}=\sum_{k' =1}^{k-1} \sum_{b'> b_{k'}} \sum_{i\in S_{b'}^{k'}}  w_iy_i + \sum_{b'\geq b}\sum_{i\in S_{b'}^{k}}  w_iy_i 
\end{equation}
\begin{equation}\label{eq:cumsumW}
W_{b}^{k}=\sum_{k' =1}^{k-1} \sum_{b'> b_{k'}} \sum_{i\in S_{b'}^{k'}}  w_i^2 + \sum_{b'\geq b}\sum_{i\in S_{b'}^{k}}  w_i^2 
\end{equation}
\begin{equation}\label{eq:cumsumN}
N_{b}^{k}=\sum_{k' =1}^{k-1} \sum_{b'> b_{k'}} \sum_{i\in S_{b'}^{k'}}  1 + \sum_{b'\geq b}\sum_{i\in S_{b'}^{k}}  1
\end{equation}
are cumulative sums of all the buckets discarded because we know they belong to $S*$ from previous iterations, and all the larger or equal buckets of the iteration $k$. 
Both of these values will be used for incrementally processing $\varrho_j(w,y)$, equation (\ref{eq:varrhok-w}).
More precisely, we define:
\begin{equation}
\varrho_{N_b^{k}}(w,y)=(C_{b}^{k}-a)/W_b^{k}.
\end{equation}
Let $b$ be the largest value such that $\bucz_b^k$ is not empty. If $\varrho_{N_b^{k}}(w,y) \geq \min \bucz_b^k$, then $b_k = b$ and we can discard all the buckets with $b' < b$ and continue to the next iteration, since $\lambda \geq \varrho_{N_b^{k}}(w,y)$.
Otherwise, if $\varrho_{N_b^{k}}(w,y) < \min \bucz_b^k$, then $S_b^k \subseteq S^*$ and we can continue processing the other buckets. 
Let $b_k$ be the largest value such that $\bucz_{b_k}^k$ is not empty and $\varrho_{N_{b_k}^{k}}(w,y) \geq \min \bucz_{b_k}^k$. 
We know that for all $b>b_k$, $S_b^k \subseteq S^*$, and that for all $b<b_k$, $\forall v \in \bucz_b^k, v < \varrho_{N_{b_k}^{k}}(w,y) \leq \lambda$. 
Thus we can safely go to the next iteration, considering only $\bucz_{b_k}^k$. 
Note that if $y \in \Delta_a$, then such a $b_k$ value does not exist, and at the first iteration we can stop.

From the definition of $I^k$ (\ref{eq:Ibound}), we can show that the size of the bucket $\bucz_{b_k}^k$ is strictly decreasing. 
Let $\bar{k}$ be the iteration where $z_{(K)}$ is the minimum value of a bucket $\bucz_{b_{\bar{k}}}^{\bar{k}}$. 
Let $b'$ be the largest value, strictly lower than $b_{\bar{k}}$, such that $\bucz_{b'}^{\bar{k}}$ is not empty. 
We have:
\begin{equation}
    \max \bucz_{b'}^{\bar{k}} < \varrho_{N_{b_{\bar{k}}}^{\kmax}}(w,y)
    \wedge \min\bucz_{b_{\bar{k}}}^{\kmax} \geq \varrho_{N_{b_{\bar{k}}}^{\kmax}}(w,y).
\end{equation}
Such a condition, from equation (\ref{eq:K-w}), implies that $\varrho_{N_{b_{\bar{k}}}^{\kmax}}(w,y) = \lambda$. 
The complexity of the \wbuckP algorithm is highly dependent of the bucketing function $h^{k}$, and using equation (\ref{eq:Ibound}), we can easily show that the worst case complexity is bound by $O(d^2)$. 
Following the same idea as \cite{perez19}, we can use a bucketing function based on the numbers encoding in nowadays computers. 
Such a function, at each iteration, choose to partition the numbers with respect to their $k$th byte, in the same fashion as the Radix sort \cite{thomas2001}.
Using such a bucketing function loose the property of equation (\ref{eq:Ibound}), but the advantage is that the complexity becomes linear $O(d+B)$.
An implementation is given in Algorithm \ref{a:bucket}.

\begin{algorithm}[t]
\normalsize
\KwData{$y, w, a$}
$\bucy_{b_0}^0 \leftarrow y$ \\
$C_{b_0}^0 \leftarrow -a$ \\
$W_{b_0}^0 \leftarrow 0$\\
\nl \For{$k \in 1..\ceil*{\log_b(D)}$}{
    \For{$b \in 1..B$}{
	    $S_b^{k} \leftarrow \{i\in S_{b_{k-1}}^{k-1}\;|\;h^{k-1}(y_i)=b\}$\\
	    $\bucy_b^{k} \leftarrow (y_i)_{i\in S_b^{k}}$
	}
	\nl \For{$b \in B..1$}{
	    $b_k \leftarrow b$\\
		\uIf{$\varrho_{N_{b+1}^{k}}(w,y) > max(\bucy_{b}^{k})$}{
			break loop \textbf{1}  
		}
		\If{$\varrho_{N_{b}^{k}}(w,y) \geq min(\bucy_{b}^{k})$}{
			break loop \textbf{2}
		}
	}
}
$\tau \leftarrow \varrho_{N_{b_k}^{k}}(y)$\\
\For{$i \in 1..|y|$}{
	$x_i \leftarrow \max(y_i - w_i\tau, 0)$
}
\caption{\wbuckPn \label{a:bucket}}
\end{algorithm}

\textbf{Filtering} 
The advantage of algorithm \wcondP is to discard values that are known to be already dominated by an incrementally updated lower bound of $\lambda$. 
In \wbuckP, we can easily use the exact same lower bound, by keeping its process in parallel of the processing of the buckets. 
Moreover, thanks to the bucketization, we can also have another lower bound. 
When we are processing bucket $b$ at iteration $k$, then $\varrho_{N_{b+1}^{k}}(w,y)$ is another, pretty good lower bound of $\lambda$. 
Note that this value should be directly available, because it was required to process the previous bucket. 
Finally, the filtering consists in removing values that are lower than one of our lower bounds.

\section{Experimental evaluation}\label{sec:experiments}
In these experiments, we reproduced the experiments from \cite{perez19} by defining random vectors of size varying between $10^5$ and $10^7$, using
either uniform or Gaussian distributions. 
We generated 500 vectors for each experiment and ran each algorithm independently, and extracted their mean times. 
We show here the performances of the proposed algorithms, \wcondP, and \wbuckPn and \wbuckP against the existing algorithm $w$-$sort$ \cite{slavakis10} Algorithm \ref{a:sort}, implemented using an efficient quick-sort procedure. The difference between \wbuckPn and \wbuckP is the use or not of the filtering improvement.
All the algorithms are implemented in C, and run on a I7 3rd generation. All source codes are available online\footnote{Link hidden}.

\begin{figure}[]
    \centering
    \includegraphics[width=\figsize]{plot/all_sizes_a4_rand0_sigma0-crop.pdf}
    \includegraphics[width=\figsize]{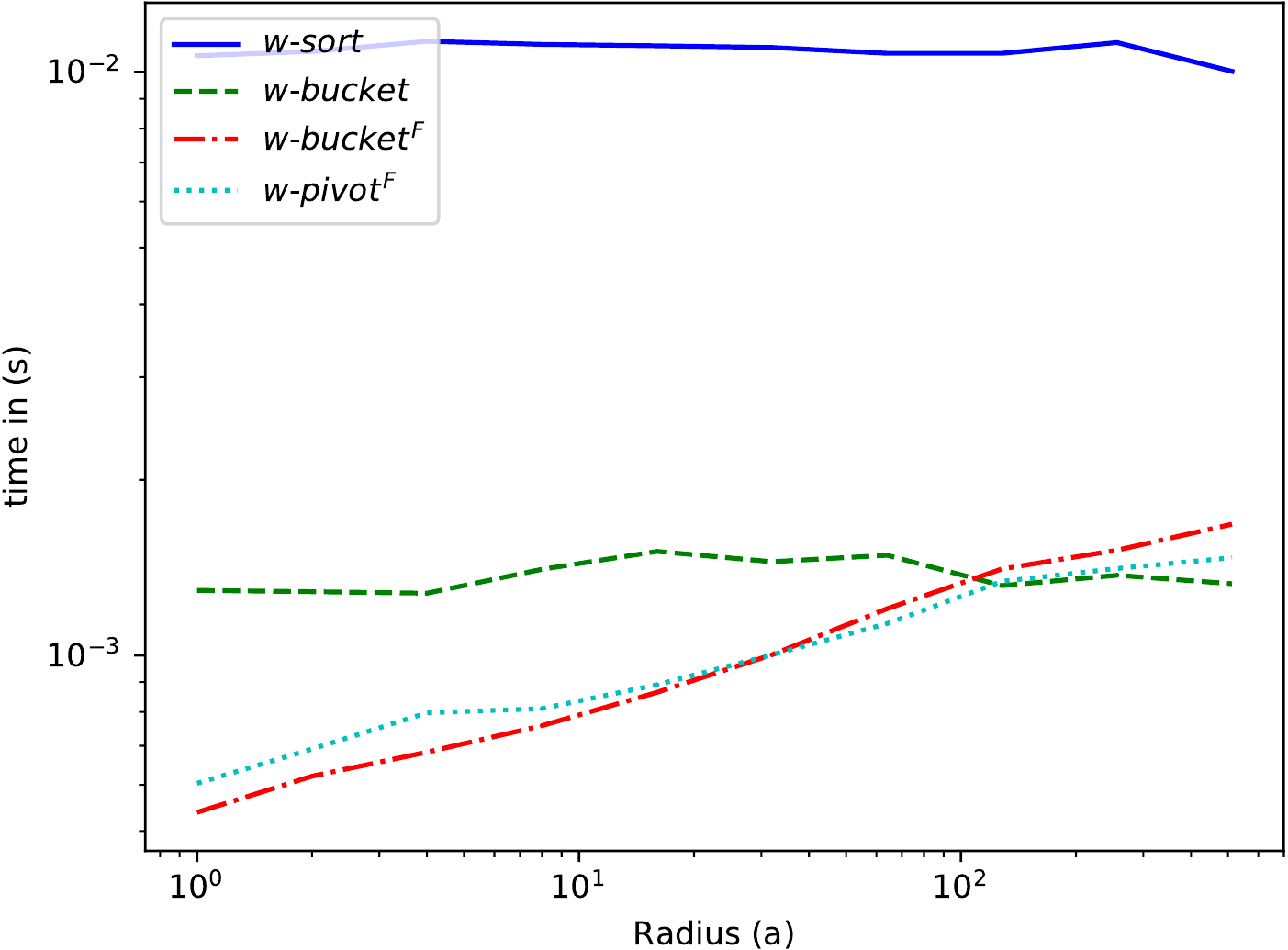}
    \caption{Uniform distribution - Top:  Projection time comparison, while the $d$ value (size of the vector $y$ to project) changes from $10^5$ to $10^7$, with $a=4$. Bottom: Projection time comparison, while the radius $a$ changes from $1$ to $512$, with $d=1O^5$.}
    \label{fig:uniform}
\end{figure}

\textbf{Uniform}
We start our experiments with Figure~\ref{fig:uniform}. 
This figure shows that when a uniform random distribution is used for vector $y$, the time needed for projecting the vector grows linearly for all methods as a function of the vector size.
Moreover, the proposed algorithms seem to perform order of magnitude faster than $w$-$sort$, which is already a faster algorithm than the state of the art. 
The second plot of Figure~\ref{fig:uniform} shows the impact of the radius for the projection time. 
It is interesting to note that while all proposed methods outperform the sorting scheme, when the radius becomes too large, the cost of filtering become larger than the gain it can bring, because less values are discarded.

\begin{figure}
    \centering
    \includegraphics[width=\figsize]{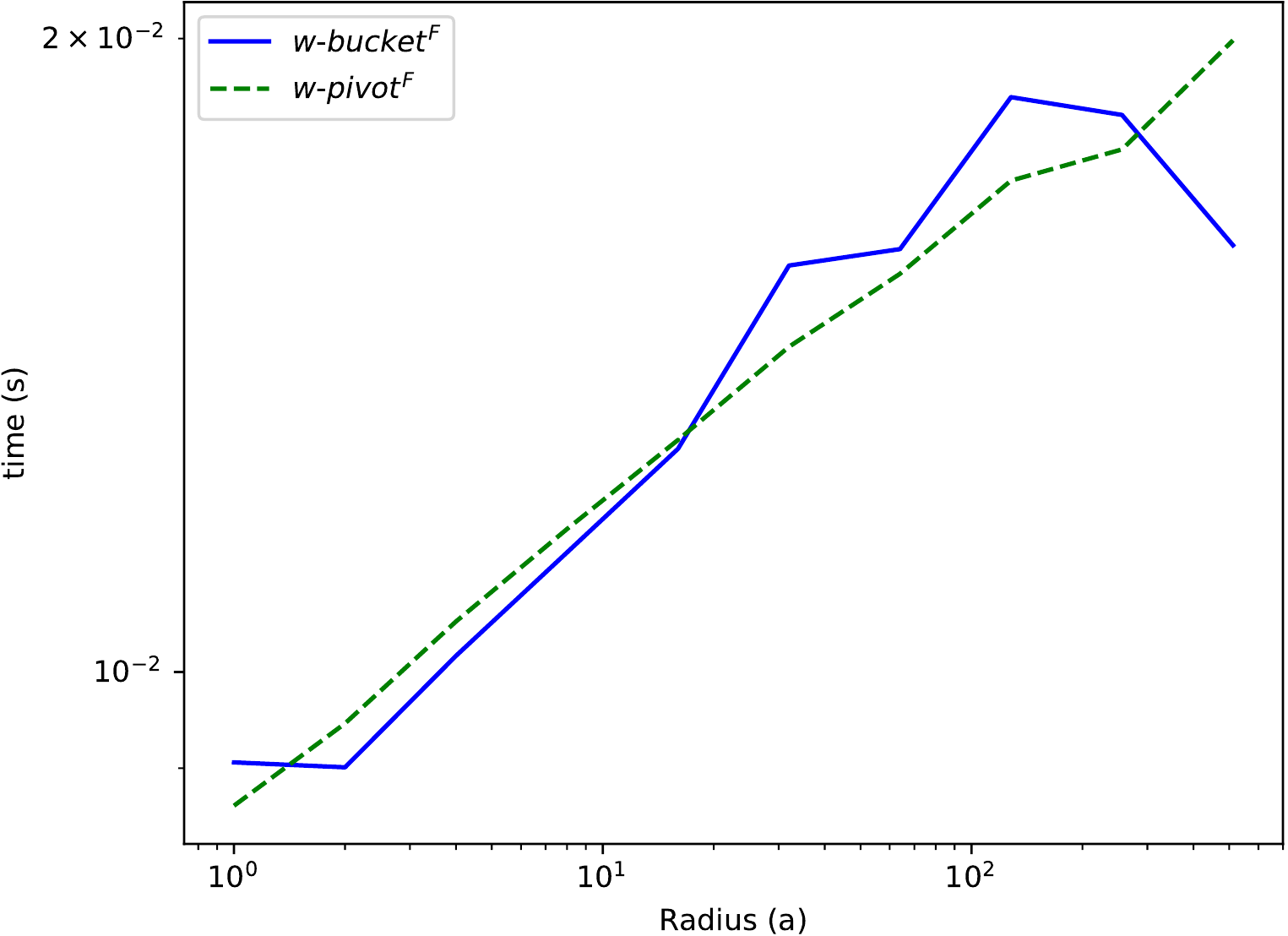}
    \includegraphics[width=\figsize]{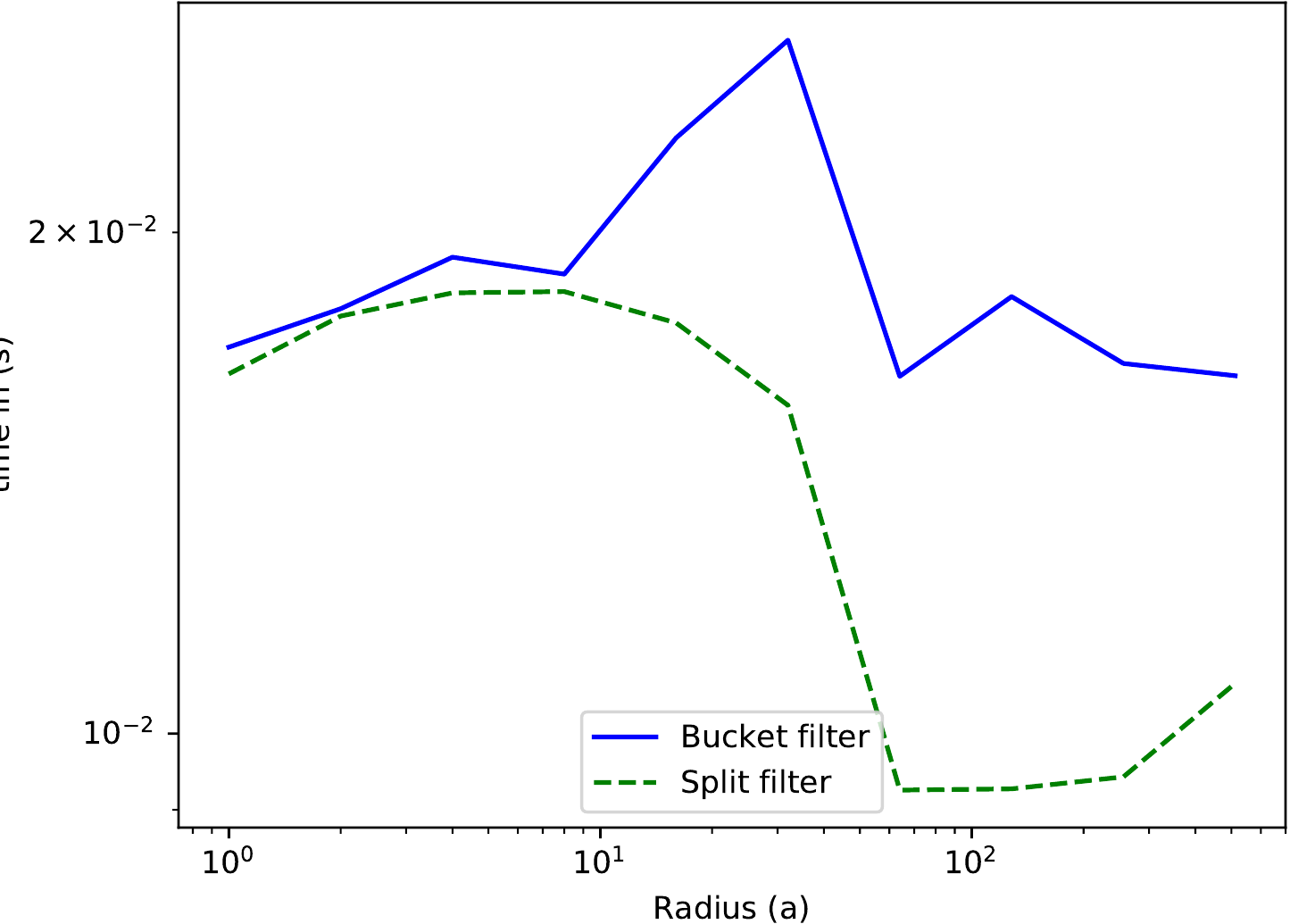}
    \includegraphics[width=\figsize]{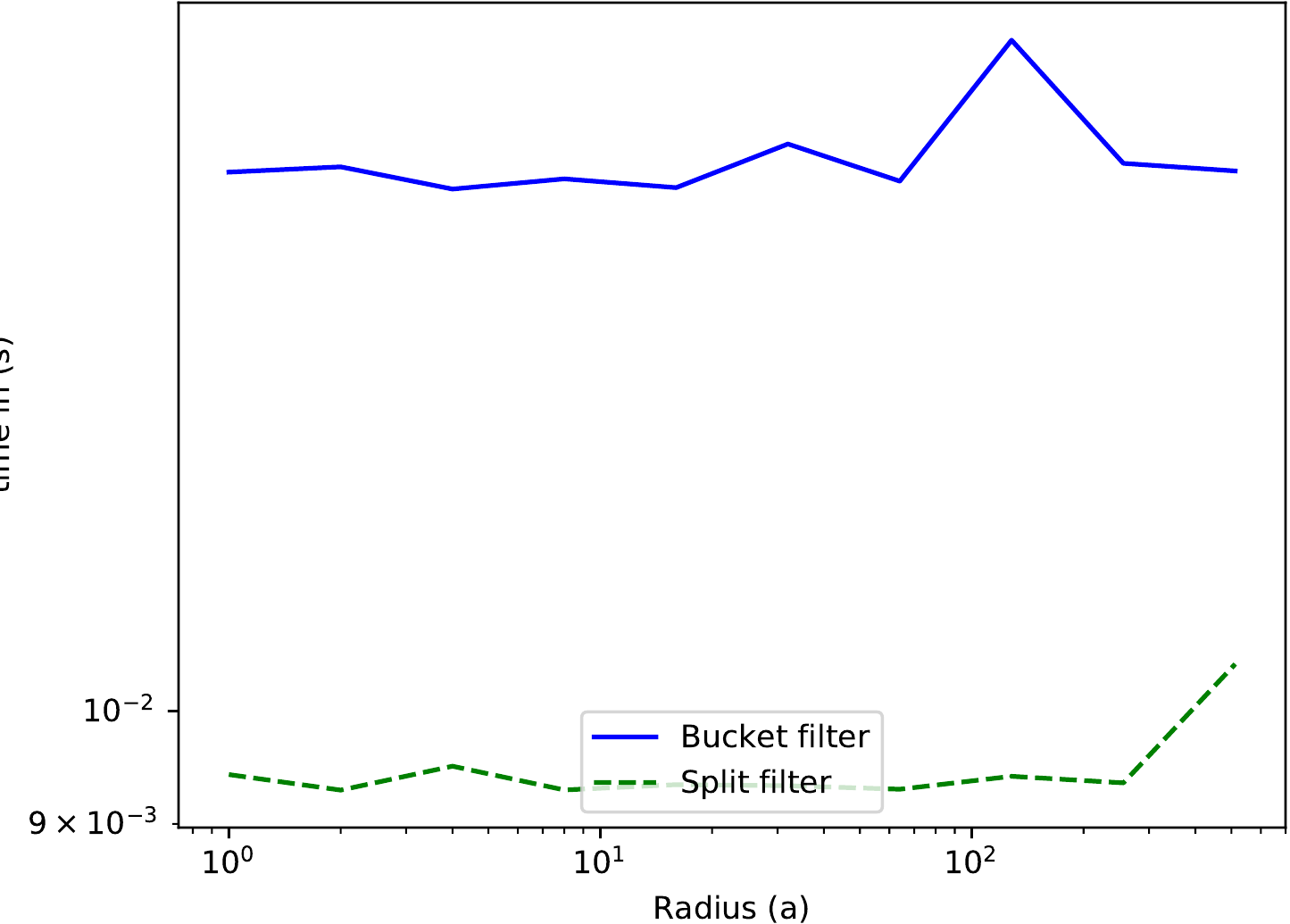}
    \caption{Gaussian law, impact of std-dev. All three experiments have $d=10^6$. Top: std-dev=$10^{-1}$. Middle: std-dev=$10^{-2}$. Bottom: std-dev=$10^{-3}$ }
    \label{fig:gaussian_a_val_filter}
\end{figure}

\begin{figure}[]
    \centering
    \includegraphics[width=\figsize]{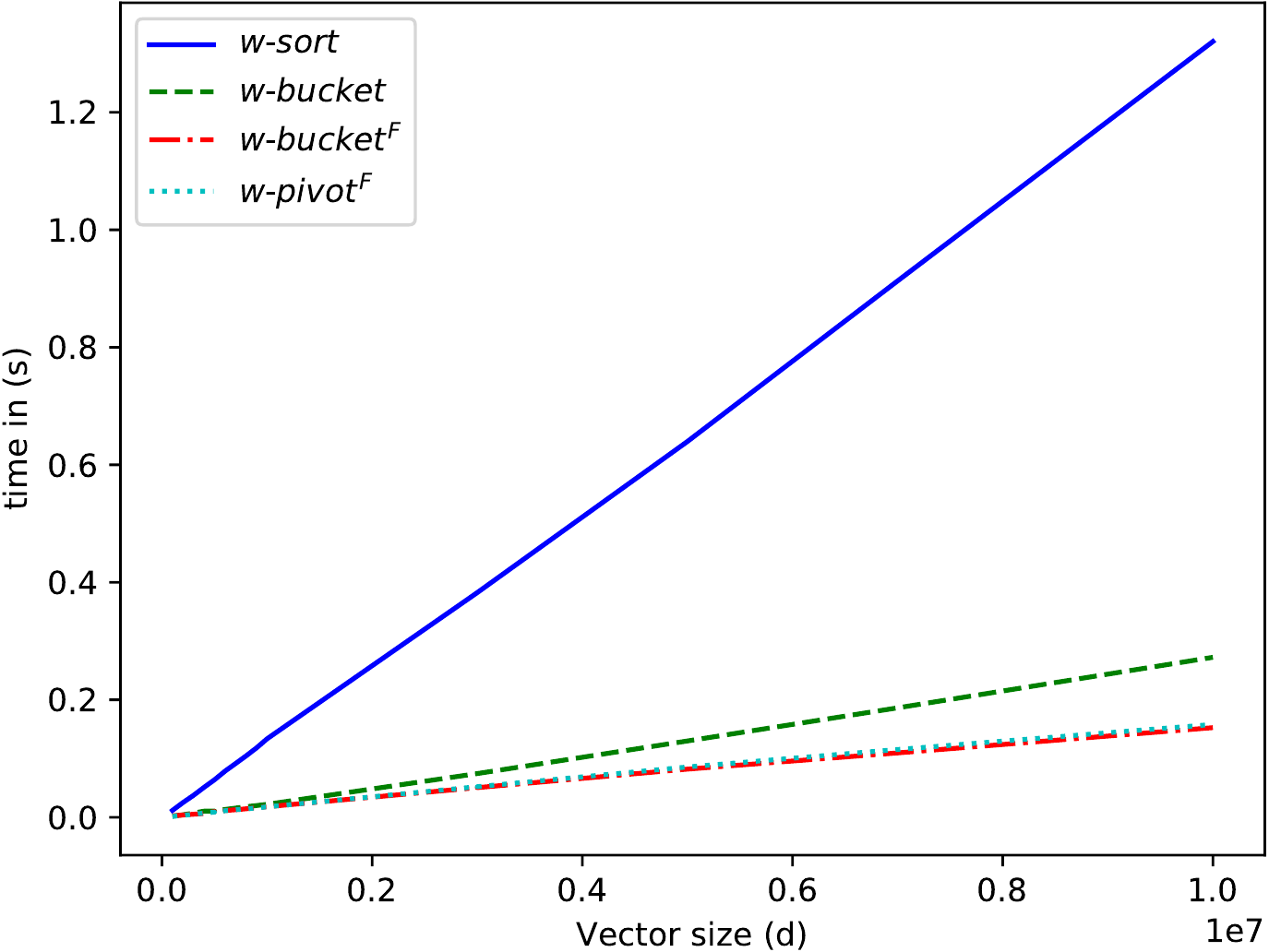}
    \includegraphics[width=\figsize]{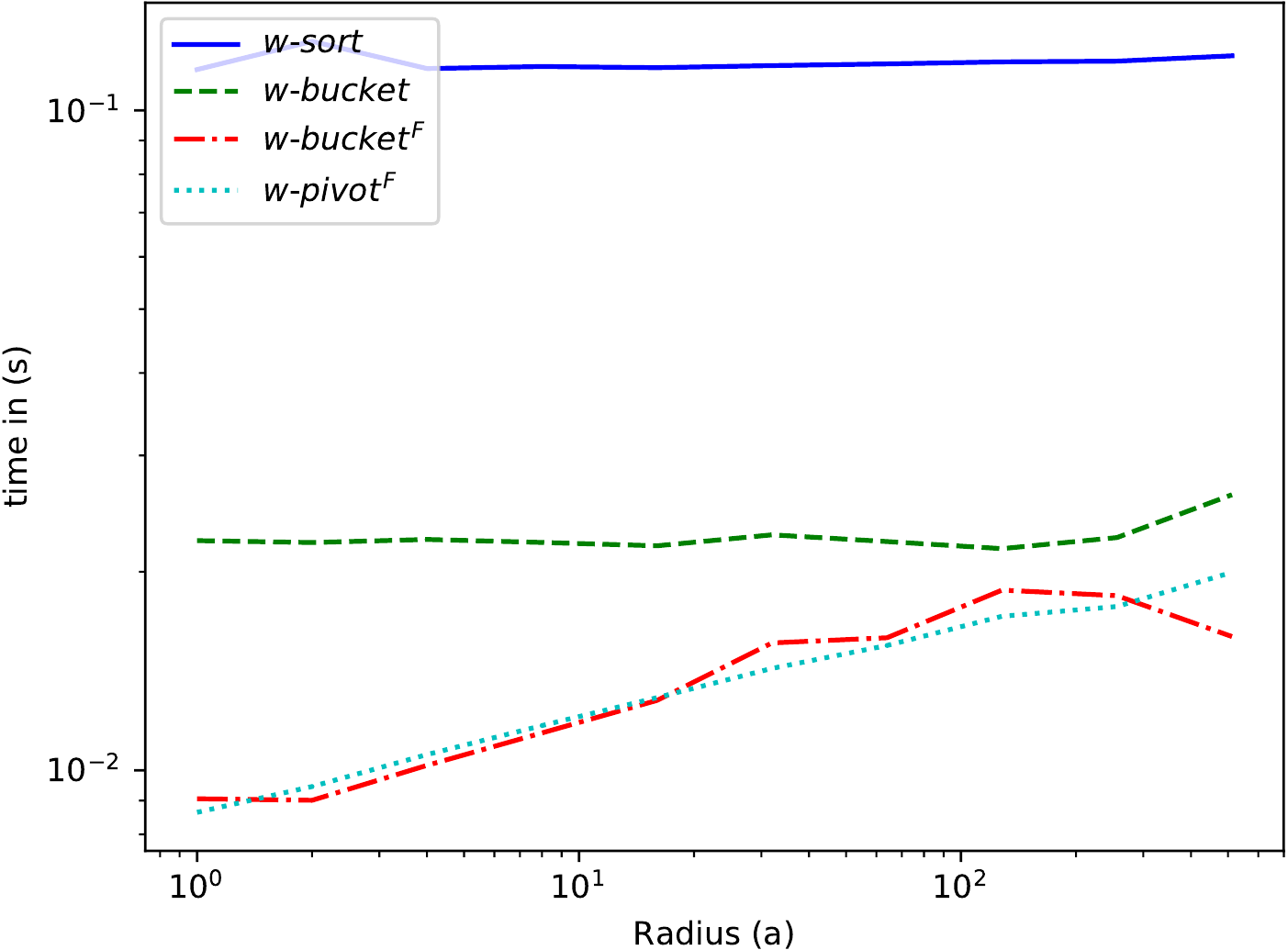}
    \caption{Gaussian law:Top: Projection time comparison, while the $a$ value changes from 1 to 512, with $d=10^6$ and std-dev = $10^{-1}$.
     Bottom:  Projection time comparison, while the $d$ value changes from $10^5$ to $10^7$, with $a=1$ and std-dev = $10^{-2}$. }
    \label{fig:gaussian_a_val}
\end{figure}

\textbf{Gaussian}
Figure~\ref{fig:gaussian_a_val} (top) shows that the time seems to grow linearly with $d$, the size of vector $y$. 
Moreover, the \wbuckP and \wcondP algorithms perform better. 
Figure~\ref{fig:gaussian_a_val} (bottom) shows that first, when the radius is small or unit, the filtering algorithms are the most efficient, but the more the radius grows, the less they are, and the classical bucket become the best one. 
Such a result may imply that in function of the radius size, one should choose to use the filtering or not in the \wbuckPn algorithms. 
Figures~\ref{fig:gaussian_a_val_filter} shows the different results for the filtered algorithms only, since their running times are order of magnitude faster. 
As we can see, they behave similarly in most of the cases, which may imply that the larger cuts in the search are induced by the filtering scheme, rather than the splitting scheme.

\textbf{Discussion}
Differences in terms of running time between \wbuckP and \wcondP are relatively small in practice. 
From a theoretical point of view, both of these algorithms take advantage of the filtering. 
Thus implementing both algorithms in a projected gradient descent solver would be useless. 
We made the choice of designing these two algorithms because they represent some of the current best state of the art algorithms for the $\ell_1$ ball.
From our point of view, only the \wbuckP algorithm should be implemented because of its efficiency and complexity guaranties.

\section{Variables Selection}
In signal reconstruction and variables selection, the non-smooth $\ell_p$ ($0 < p < 1$) 
regularization has proven efficient in finding sparse solutions.
Consider the problem of minimizing a quadratic reconstruction error subject to
the $\ell_p$ regularization, $\normP{x}=\sum_{i=1}^{d}|x_i|^p$.
\begin{equation}\label{eq:minl2lp}
\underset{x \in \RR^d}{\text{minimize}} \normtwo{Ax-b} + \lambda \normP{x}
\end{equation}
where $A \in \RR^{n\times m}$, $b \in \RR^{m}$, and $\lambda \in _{\geq0}$ is the penalty parameter.

A popular method to solve this problem is to use the iteratively reweighted  $\ell_1$ minimization (IRL1) \cite{candes2008enhancing,chartrand2008iteratively,chen2010convergence,chen2014convergence}. 
An application of IRL1 to problem (\ref{eq:minl2lp}) can be
\begin{equation}\label{eq:IRL1l2lp}
x^{k+1} \in \underset{x \in \RR^d}{\text{arg min}} \normtwo{Ax-b} \quad \text{ subject to } \quad \normone{W^k x} < r^k
\end{equation}
with the weight $W^k = \text{diag} (w^k)$, and $w^k$ is defined by the previous iterates by
\begin{equation}
    w^k_i = \frac{1}{(|x^k_i|+\epsilon)^{1-p}},\quad\quad i = 1,\dots,n.
\end{equation}
where $\epsilon\in \RR_{>0}$.

Using the weighted $\ell_1$ ball projection algorithms proposed in this paper, solving
(\ref{eq:IRL1l2lp}) can be done easily.
We propose to use the following algorithm, based on \cite{trzasko08}, we start with $x^0\in \RR^m$ randomly defined. We set $p = 1$, which is equivalent to solving the LASSO problem. 
Then, we smoothly decrease $p$, and iteratively solve an IRL1 problem such as (\ref{eq:IRL1l2lp}) with a fixed $p$. A possible implementation of this algorithm is
in Algorithm \ref{a:SIRLS}.
We call this algorithm Smooth Iterative Reweighted $\ell_1$ ball Projections (SIRL1).

\begin{algorithm}[t]
\normalsize
\KwData{$x^0 \in \RR^m,A \in \RR^{n\times m}, b \in \RR^{m}$}
$p \leftarrow 1$ 

\While{$p > 0$}{
\While{IRLS-p hasn't converged}{
$w^k_i = \frac{p}{(|x^k_i|+\epsilon)^{1-p}},\quad\quad i = 1,\dots,n.$

\begin{equation*}
x^{k+1} \in 
    \hspace{-1.75cm}\begin{split}
        \underset{x \in \RR^d}{\text{minimize}} \normtwo{Ax-b} \\
        \text{subject to} \normone{W^k x} < r^k 
    \end{split}
\end{equation*}
}
Decrease smoothly $p$
}
\caption{Smooth Iterative Reweighted $\ell_1$ ball Projections \label{a:SIRLS}}
\end{algorithm}

\paragraph{Reconstruction results} We show in the section the reconstruction efficiency of the SIRL1, which is a direct application of having an efficient projection onto the weighted L1 ball.
We reproduce here part of the experimental protocol of \cite{candes2008enhancing}.
We compare the reconstruction error and the sparsity against the state of the art LASSO algorithm and the projection based (PC) version of \cite{candes2008enhancing}. 
In this experiment, $n$ is the number of rows, $m$ the number of columns and $k$ the real number of non-zero components of the solution.
First, table \ref{tb:global} shows some results on the reconstruction of sparse vectors with sparsity of 5, 15, 30 non-zero values over 100 values. 
As we can see, The reconstruction accuracy and the sparsity is rapidly found with even a small number of different $p$. 
Moreover, because of the smoothness induced by $p$, even if the number of iterations is larger in (a=5,$\#p=5$) compared to (a=5,$\#p=3$), the running time is slower. 

\begin{figure}[t]
    \centering
    \includegraphics[width=8cm]{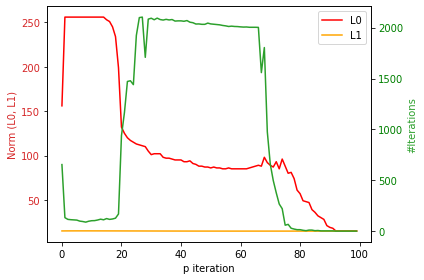}
    \caption{Relationship between the number of iterations, $p$, and the norms (n=100, m=256, k=15). }
    \label{fig:iterations}
\end{figure}

A look at the impact of the smoothness between the values of $p$ is given in Table \ref{tb:smoothness}. 
We can see that less than 3 iterations over $p$ may be too small, and that more than 5 in this example seem to be useless.
But one must be careful with the smoothness, we tried to push the smoothness to 100 iterations over $p$, the results are in Fig~\ref{fig:iterations}.
As we can see, the number of iterations between $p=0.2$ and $p=0.8$ is high, while the L0 norm is not impacted.
We were able to see this kind of results in different settings we set, with larger $n$, $m$ and $k$.
Finaly, the radius is one of the most important parameter of this algorithm, it's impact can be seen on table \ref{tb:radius}.
As it is expected, a radius smaller than the number of non-zero components has a bad impact on the reconstruction, but it is interesting to see that less iterations seem to be required to solve these problems than for larger values of the radius.
Finally, the results we obtained are coherent with the results obtained in \cite{trzasko08}, which validates the inverted model we used.

\begin{table}[h!t]
\footnotesize
\begin{tabular}{|l|l|l|l|l|l|l|}
\hline
\textbf{Algo}   & \textbf{L0}  & \textbf{L1}    & \textbf{rec}  & \textbf{time}   & \textbf{\#it} & \textbf{a}    \\
\hline
Lasso  & 5   & 5.00  & 1.00 & 0.0029 & 11   & 5.00 \\
PC     & 5   & 4.86  & 4.82 & 1.1295 & 3880 & 5.00  \\
$\#p=3$   & 5   & 5.22  & 0.00 & 0.2542 & 2107 & 5.00 \\
$\#p=5$   & 5   & 5.22  & 0.00 & 0.1844 & 2176 & 5.00  \\
Lasso  & 151 & 15.00 & 0.01 & 0.0163 & 159  & 15.00 \\
PC     & 256 & 14.80 & 0.00 & 0.9570 & 307  & 15.00 \\
$\#p=3$    & 15  & 14.80 & 0.01 & 0.3580 & 4909 & 15.00  \\
$\#p=5$    & 15  & 14.80 & 0.00 & 0.3790 & 5201 & 15.00  \\
Lasso  & 146 & 30.00 & 0.05 & 0.0223 & 240  & 30.00 \\
PC     & 256 & 27.73 & 0.00 & 0.0477 & 352  & 30.00  \\
$\#p=3$    & 30  & 27.73 & 0.00 & 0.1563 & 2053 & 30.00\\
$\#p=5$    & 30  & 27.73 & 0.00 & 0.5103 & 7131 & 30.00 \\
\hline
\end{tabular}
\caption{Reconstruction and Sparsity for various State of the art algorithm and the SIRL1. 
(n=100, m=256) \label{tb:global}}
\begin{tabular}{|l|l|l|l|l|l|l|l|l|}
\hline
\textbf{L0}  & \textbf{L1}    & \textbf{rec}  & \textbf{time}   &\textbf{ \#it}  & \textbf{a}     & \textbf{$\#p$} \\
\hline
156 & 15.00 & 0.00 & 0.0580 & 655   & 15.00 & 1  \\
15  & 14.76 & 0.15 & 0.0600 & 796   & 15.00 & 2  \\
15  & 14.80 & 0.01 & 0.3429 & 4909  & 15.00 & 3  \\
15  & 14.80 & 0.00 & 0.3770 & 5201  & 15.00 & 5 \\
15  & 14.80 & 0.00 & 0.4386 & 5929  & 15.00 & 6  \\
15  & 14.80 & 0.00 & 1.1664 & 11197 & 15.00 & 10 \\
5   & 5.00  & 1.00 & 0.0050 & 14    & 5.00  & 1  \\
5   & 5.03  & 0.91 & 0.6068 & 3032  & 5.00  & 2  \\
5   & 5.22  & 0.00 & 0.1810 & 2107  & 5.00  & 3  \\
5   & 5.22  & 0.00 & 0.1766 & 2176  & 5.00  & 5  \\
5   & 5.22  & 0.00 & 0.2325 & 2227  & 5.00  & 6  \\
5   & 5.22  & 0.00 & 0.1939 & 2446  & 5.00  & 10 \\
\hline
\end{tabular}
\caption{Impact of the smoothness of Q on the reconstruction sparsity. 
(n=100, m=256) \label{tb:smoothness}}
\begin{tabular}{|l|l|l|l|l|l|}
\hline \textbf{L0} & \textbf{L1}    & \textbf{rec}   & \textbf{time}   &\textbf{ \#it} & \textbf{Radius} \\
\hline
8  & 11.58 & 13.18 & 0.0328 & 316  & 7.50  \\
9  & 12.05 & 11.69 & 0.0265 & 247  & 8.62    \\
10 & 13.13 & 8.73  & 0.1676 & 1072 & 9.75    \\
11 & 13.33 & 7.61  & 0.0452 & 459  & 10.88   \\
12 & 13.82 & 3.34  & 0.2793 & 2251 & 12.00   \\
14 & 14.50 & 1.95  & 0.0318 & 306  & 13.12   \\
15 & 14.69 & 1.02  & 0.0427 & 528  & 14.25   \\
15 & 14.80 & 0.00  & 0.3668 & 5411 & 15.38   \\
17 & 14.80 & 0.00  & 0.4542 & 5607 & 16.50   \\
19 & 14.80 & 0.00  & 0.4216 & 5829 & 17.62   \\
20 & 14.80 & 0.00  & 0.4015 & 5601 & 18.75   \\
20 & 14.80 & 0.00  & 0.4011 & 5754 & 19.88  \\
\hline
\end{tabular}
\caption{Impact of the radius on the reconstruction sparsity. 
(n=100, m=256, k=15, $\#p=4$) \label{tb:radius}}
\end{table}

\section{Conclusion}
Data and feature sizes are ever increasing in nowadays problems.
In this paper we proposed 3 efficient projection algorithms with different complexities, including linear time, for working with very large problems.
Two of the proposed algorithms project very large non-sparse vectors in a small amount of time, such as 8 ms for vectors of size $10^7$, and seem to be robust to the randomness of the vector.
We showed how to directly use them in basic sparse-vector reconstruction frameworks and obtain state of the art results.
Finally, we empirically proved that they should be used as a basis for projected gradient descent frameworks working with $\ell_1$ balls, weighted or not.

\section{Proofs}
\textbf{Proof of existence of $\lambda^*$}
The projection onto the weighted simplex can be formulated as:
\begin{eqnarray*}
&\arg \min \limits_{x} &  \normtwo{x-y} \\
&\text{subject to }&  wx =a \\
& &    x_i \geq 0, \forall i \in [0,d]
\end{eqnarray*}
Whose Lagrange dual is:
$\arg \min\limits_{x} \normtwo{x-y} + \lambda (wx-a) + \mu x$
Using the Kuhn-Tucker theorem \cite{hanson1981sufficiency}, we can show that necessary and sufficient conditions for $x$ to be a optimum are
$x_i - y_i = \mu_i - w_i\lambda$, $\mu_i \geq 0,$ and $\mu_i x_i = 0$.
%
If we define $x$ and $\mu$ to be respectively 
$x_i   = \max( y_i - w_i\lambda,0), \mu_i = \max( w_i\lambda - y_i, 0)$
then the conditions are respected.
From this definition, it follows that $x_i \geq 0$ for all $i$.
Let $C(\lambda) = \sum_i w_i \max( y_i - w_i\lambda,0) = \sum_i w_i x_i$.
Our goal is to find the value of $\lambda^*$ such that $C(\lambda^*) = a$.

\begin{figure}
    \centering
    \includegraphics[width=6cm]{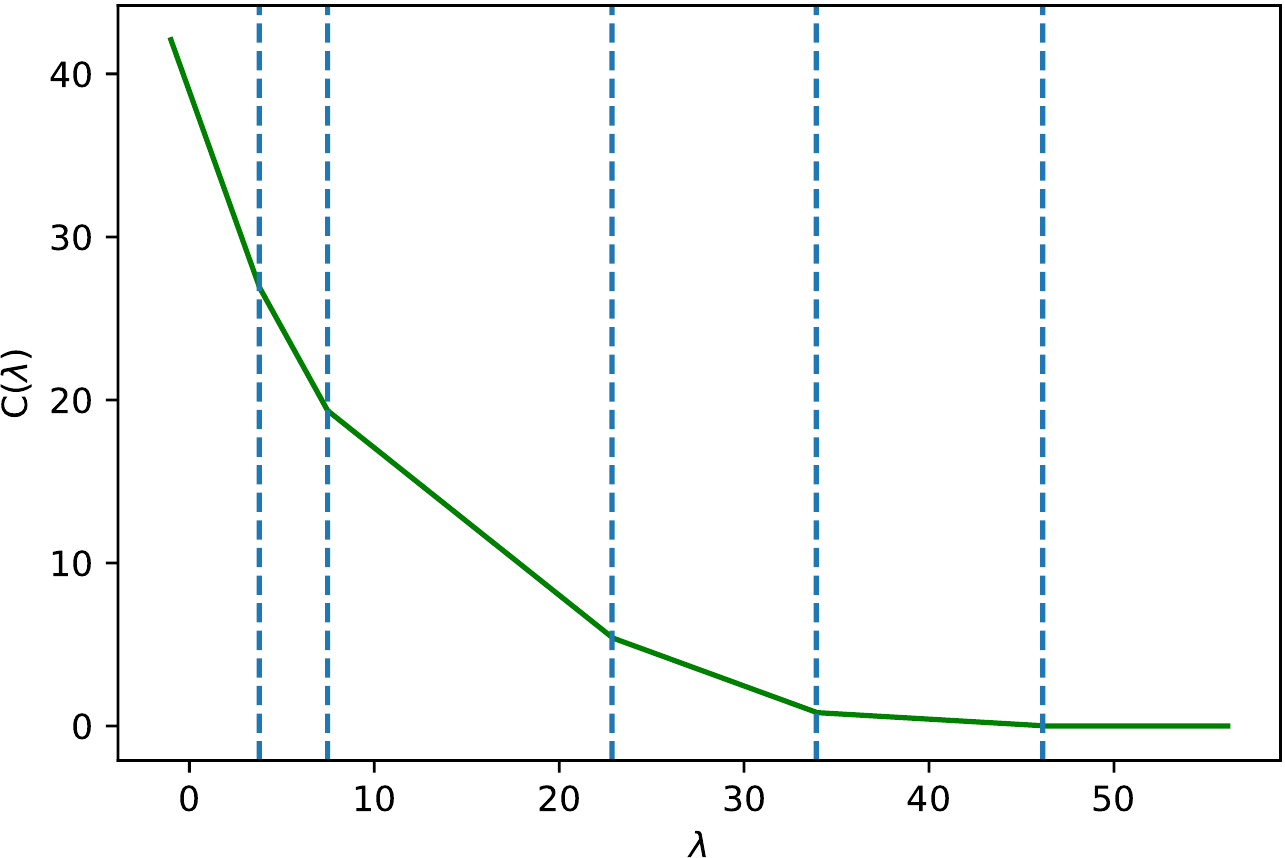}
    \caption{Example of C function. Looking for a projection onto $a=10$ is equivalent to looking for $C(\lambda)=10$. Each blue vertical line is the position of a $z_i$ value.}
    \label{fig:Cfunction}
\end{figure}

We know that the $C$ function cannot be negative and can reach any positive value.
That $C$ is piece-wise linear, and that the pieces of linearity are delimited by the values of $z = \{ \frac{y_i}{w_i} | \forall i \in [0,d] \}$.
Moreover, $C$ is non-increasing. 

Suppose $z$ sorted in increasing order, thus $z_1 \leq z_2 \leq ... \leq z_d$, let $()$ denote the permutation of $y$ and $w$ to $z$ such that $\frac{y_{(1)}}{w_{(1)}} \leq \frac{y_{(2)}}{w_{(2)}} \leq ... \leq \frac{y_{(d)}}{w_{(d)}}$.
The values at the vertices of $C$ are 
\begin{eqnarray*}
    C(z_i) &=& \sum_{j=1}^d w_j \max(y_j -\frac{w_j y_{(i)}}{w_{(i)}}, 0)\\
    C(z_i) &=& \sum_{j=i+1}^d w_{(j)} (y_{(j)} -\frac{w_{(j)} y_{(i)}}{w_{(i)}})\\
    C(z_i) &=& \sum_{j=i+1}^d w_{(j)} y_{(j)} - \sum_{j=i+1}^d\frac{w_{(j)}^2 y_{(i)}}{w_{(i)}}\\
\end{eqnarray*}

Since $C(\lambda^*) = a$, then for any $z_i < \lambda^*$, $x_i = 0$ and $C(z_i) > a$. Thus $x_i = 0$ for $i \in [1,J]$, with:
\begin{equation}\label{eq:J}
    J := \max \left \{ j \bigg | \frac{-a+\sum_{i=j+1}^d w_{(i)} y_{(i)}}{\sum_{i=j+1}^d w_{(i)}^2} > z_{(j)}  
    \right \}
\end{equation}
Using $J$, w can finally get $\lambda^*$.
\begin{eqnarray*}
    C(\lambda^*) & = & \sum_{j=J+1}^d w_{(j)} y_{(j)} - \sum_{j=J+1}^d w_{(j)}^2 \lambda^* = a\\
    -\lambda^* \sum_{j=J+1}^d w_{(j)}^2   &=& - \sum_{j=J+1}^d w_{(j)} y_{(j)}+ a\\
     \lambda^*   &=& \frac{-a + \sum_{j=J+1}^d w_{(j)} y_{(j)}}{\sum_{j=J+1}^d w_{(j)}^2 }
\end{eqnarray*}
This proof shows that once $z$ is sorted, finding $J$ and $\lambda^*$ can be done in worst case linear time, as for the non-weighted version. Only one iteration gives $J$.


\textbf{Proof that each subset is a lower-bound}
Consider $V$ to be any sub-sequence of $y$. We can compute the following pivot:
\begin{equation*}
    p_V = \frac{-a + \sum_{i\in V} w_{i} y_{i}}{\sum_{i\in V} w_{i}^2} 
\end{equation*}
Then we have:
\begin{eqnarray*}
    a & = & - p_V \sum_{i\in V} w_{i}^2 + \sum_{i\in V} w_{i} y_{i} \\
    a & = & \sum_{i\in V} w_{i} (y_{i} - w_{i} p_V) \\
    a & \leq & \sum_{i\in 1..d} w_{i} \max(y_{i} - w_{i} p_V,0) 
\end{eqnarray*}

\begin{figure}
    \centering
    \includegraphics[width=\figsize]{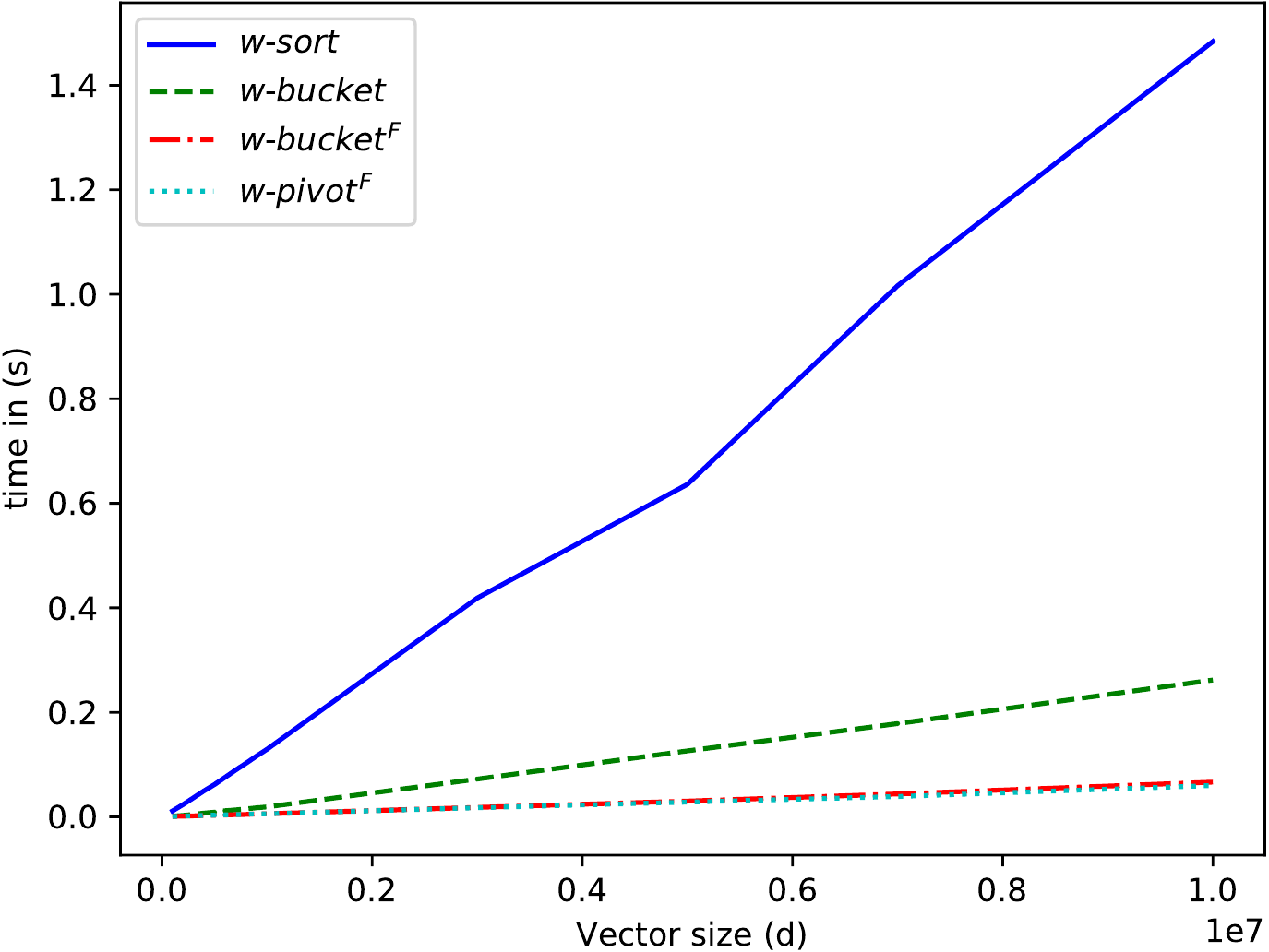}
    \includegraphics[width=\figsize]{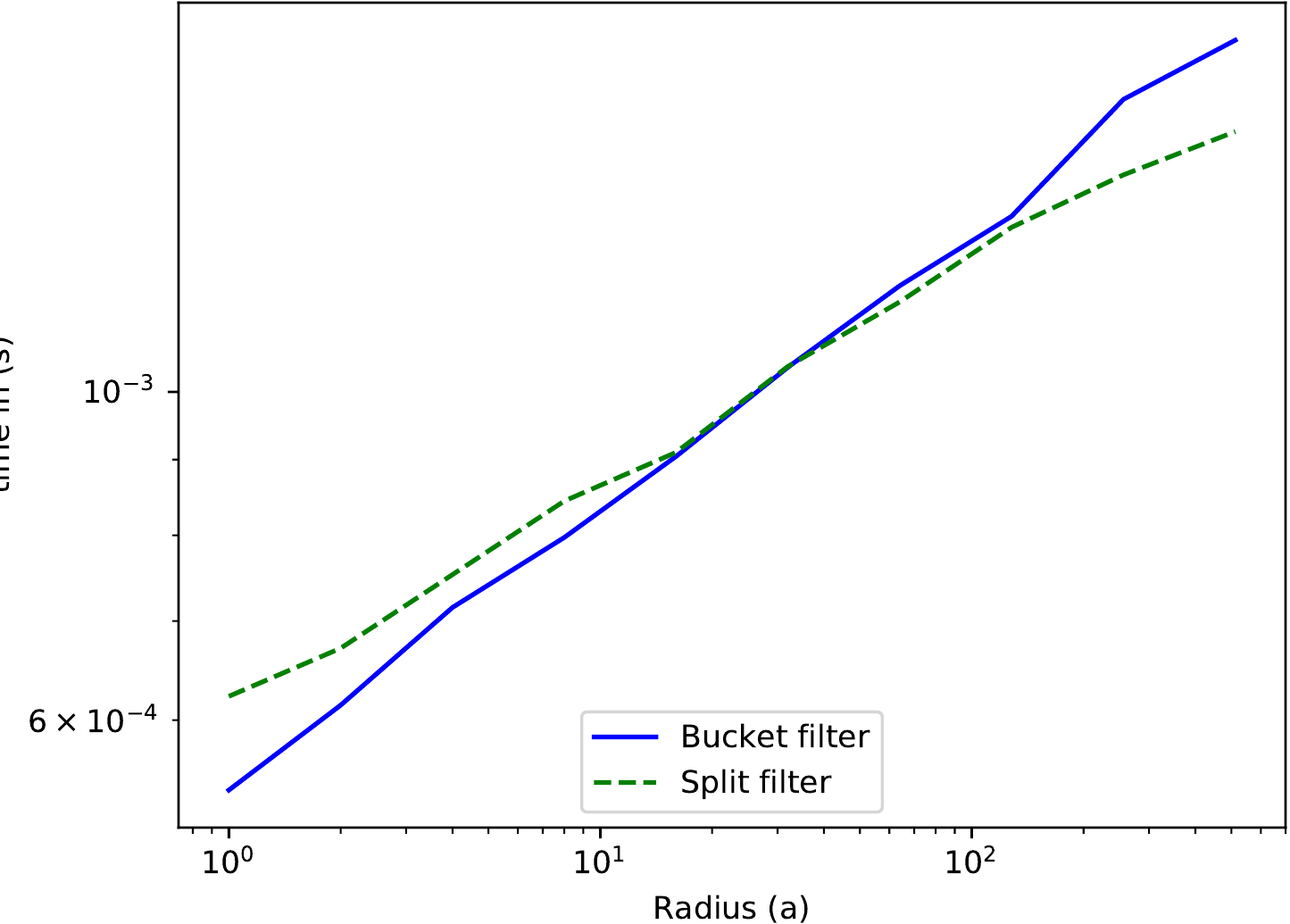}
    \caption{Uniform distribution - Top:  Projection time comparison, while the $d$ value (size of the vector $y$ to project) changes from $10^5$ to $10^7$, with $a=4$. Bottom: Projection time comparison, while the radius $a$ changes from $1$ to $512$, with $d=1O^5$.}
    \label{fig:uniform_filter}
\end{figure}

\begin{figure}[]
    \centering
    \includegraphics[width=\figsize]{plot/As_size1000000_rand1_sigma0-crop.pdf}
     \includegraphics[width=\figsize]{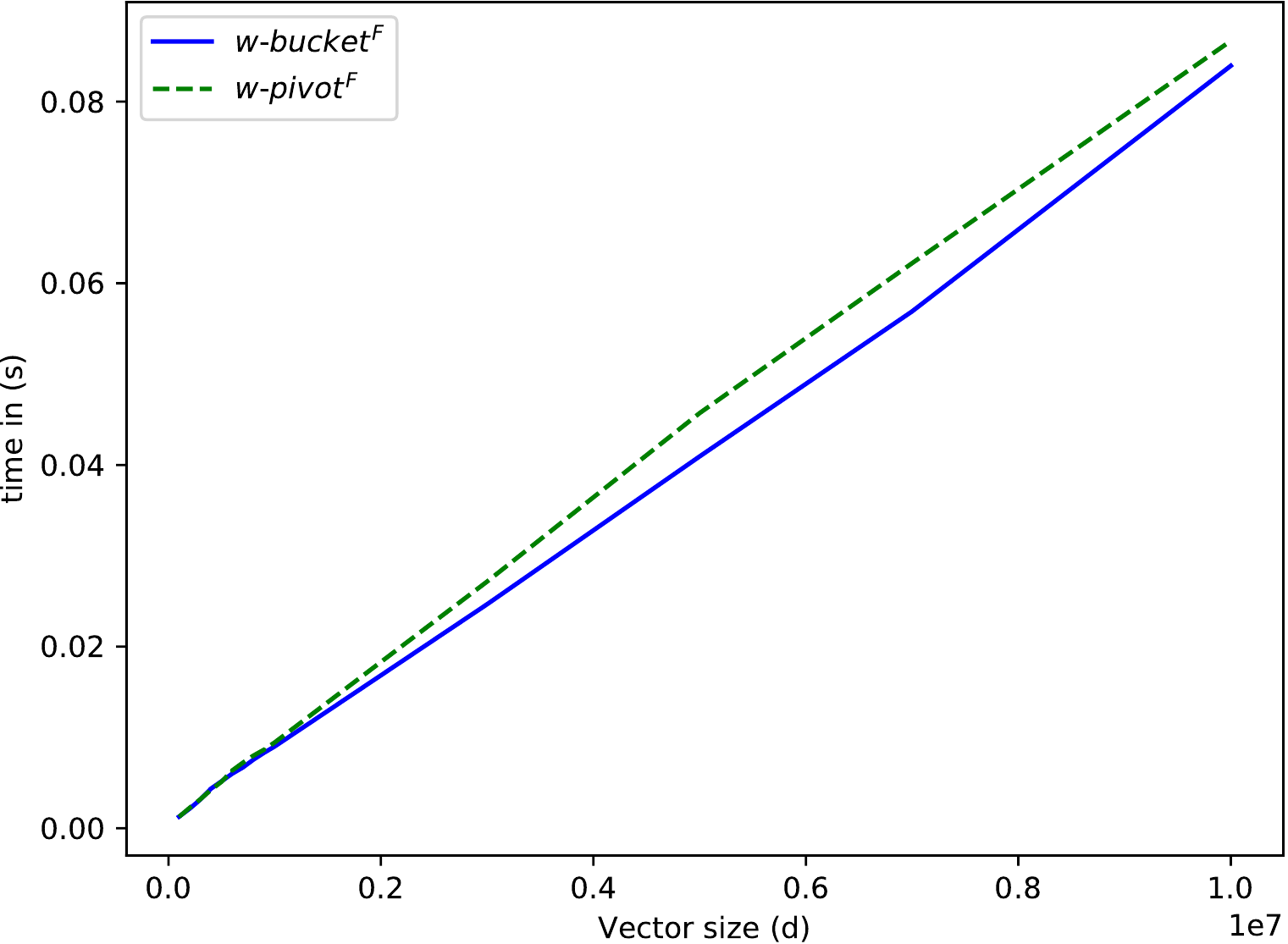}
    \caption{Gaussian law comparison of \wbuckP and \wcondP. Top: std-dev=$10^{-1}$ and $d=10^6$ 
    Bottom $a=2$ and std-dev=$10^{-1}$. }
    \label{fig:gaussian_sizes_filter}  
\end{figure}

\bibliography{gretsi}
\bibliographystyle{spmpsci}      

\end{document}